\documentclass[sigconf]{acmart}

\renewcommand\footnotetextcopyrightpermission[1]{} 
\usepackage{booktabs} 

\usepackage[utf8]{inputenc} 
\usepackage[T1]{fontenc}    
\usepackage{hyperref}       
\usepackage{url}            
\usepackage{booktabs}       
\usepackage{amsfonts}       
\usepackage{nicefrac}       
\usepackage{microtype}      
\usepackage{algorithm}
\usepackage[noend]{algpseudocode}
\usepackage{graphicx}
\usepackage{subfig}
\usepackage{bm}
\usepackage{multirow}
\usepackage{dsfont}
\usepackage{wrapfig}
\usepackage[T1]{fontenc}
\usepackage{lmodern}

\begin{document}
\title{Edge-Host Partitioning of Deep Neural Networks with Feature Space Encoding for Resource-Constrained Internet-of-Things Platforms}

\author{Jong Hwan Ko, Taesik Na, Mohammad Faisal Amir, and Saibal Mukhopadhyay}
    \affiliation{%
        \institution{School of Electrical and Computer Engineering\\
        Georgia Institute of Technology}
 \texttt{\{jonghwan.ko, taesik.na, mfamir\}@gatech.edu, saibal@ece.gatech.edu
    }}

\begin{abstract}
This paper introduces partitioning an inference task of a deep neural network between an edge and a host platform in the IoT environment. We present a DNN as an encoding pipeline, and propose to transmit the output feature space of an intermediate layer to the host. The lossless or lossy encoding of the feature space is proposed to enhance the maximum input rate supported by the edge platform and/or reduce the energy of the edge platform. Simulation results show that partitioning a DNN at the end of convolutional (feature extraction) layers coupled with feature space encoding enables significant improvement in the energy-efficiency and throughput over the baseline configurations that perform the entire inference at the edge or at the host.
\end{abstract}

%
%

\keywords{Deep learning, neural network, inference, edge, host, encoding}

\maketitle


\section{Introduction}
The growing proliferation of Internet of Things (IoT) will result in a large number of edge devices with high quality sensors streaming information to the cloud at a very high data rate. An important example is a camera network with edge devices composed of an image sensor and a lightweight processor. As deep learning techniques has shown powerful performance in intelligent processing of visual data, the integration of deep neural networks (DNNs) can greatly enhance the capabilities of such a network of edge devices.  

Typically, there are two different approaches to leveraging the deep learning capabilities in the IoT environment. 
First, the edge platforms can be designed to run deep learning inference independently, so that it can provide a local feedback or decision directly at the edge, such as controlling sensor parameters. 
More general application model is that the edge platforms are connected to the host platform, where users make decisions based on the information processed through a neural network (e.g., video surveillance \cite{vehicle}, remote monitoring \cite{monitoring}). 
This paper focuses on this edge-host combined environment, aiming to make the information from neural network inference available at the host, using limited resources at the edge devices. 
At the same time, the information should be delivered with as high throughput as possible under a bandwidth constraint, to enable better decision at the host.

One configuration to achieve this goal is to utilize the edge platform as an image source that provides the visual data to the host so that it can perform neural network inference with relatively sufficient resources [Figure 1(a)].
The challenge is to achieve high image throughput with a limited transmission bandwidth. Therefore, the input image space is normally compressed (e.g. Motion JPEG) to reduce the bandwidth demand and transmission energy dissipation at the edge.  
To avoid bandwidth-intensive transmission between the edge and the host, an edge platform can embed a neural network inference engine to process the image data directly on device [Figure 1(b)]. By transmitting the end output of inference, the transmission demand can be significantly reduced. 
However, neural network inference is a costly operation that requires large memory and computation, which degrade the energy-efficiency and throughput of the edge platforms. Moreover, solving complex problems requires deeper networks and a large number of parameters, significantly increasing the compute/memory demand at the edge device. 
\begin{figure}[t]
 \centering
 \centerline{\includegraphics[trim={6cm 2.5cm 6cm 6.8cm},clip, width=0.99\linewidth]{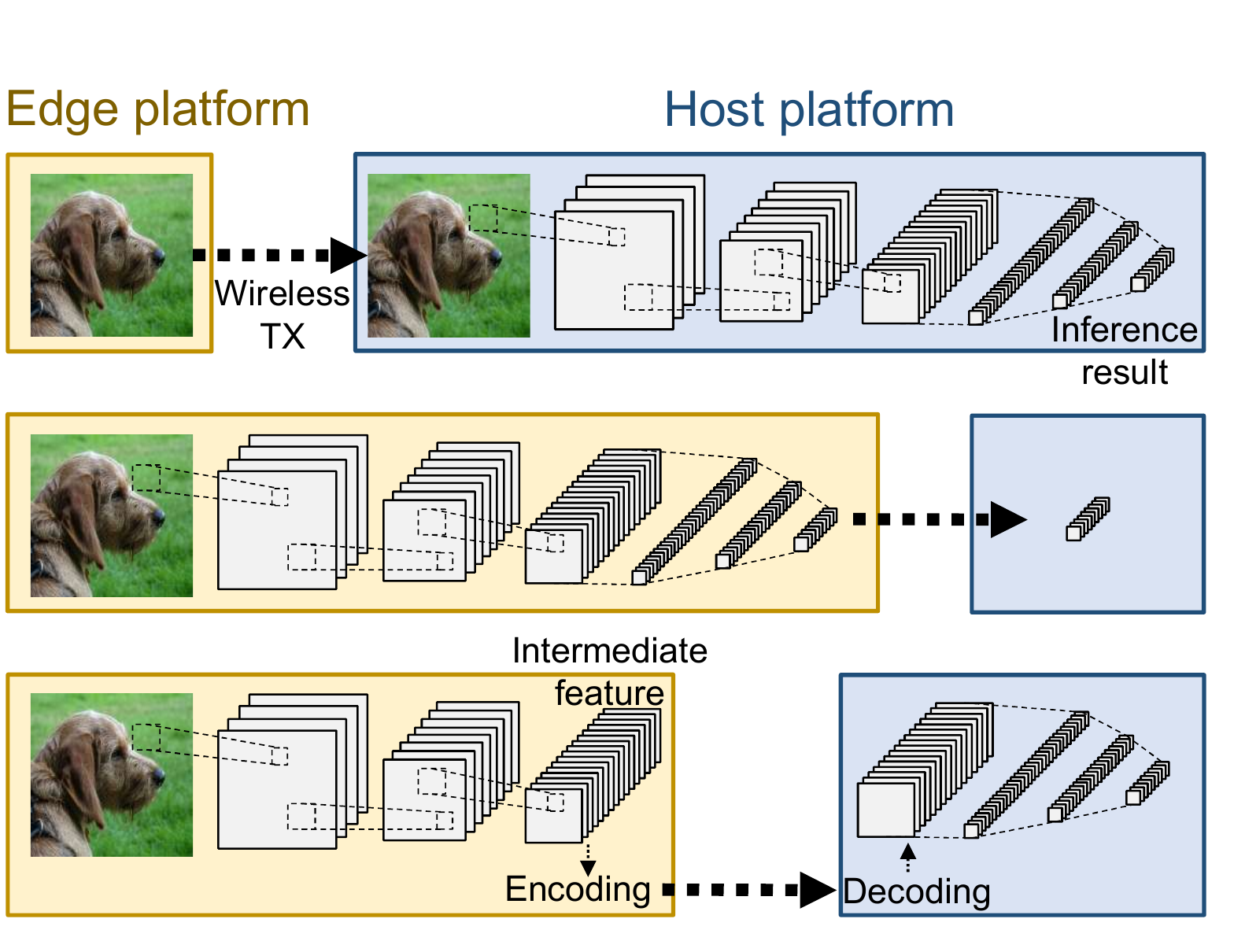}}
 \caption{ Three approaches to deep-learning inference in IoT environment with an edge and a host platform. (top) Entire inference at the host using images transmitted by the edge, (middle) entire inference at the edge that transmits the end output to the host, and (bottom) the proposed partitioned inference where the edge processes inference up to an intermediate layer, whose features are encoded and transmitted to the host for inference of the rest of the network. 
}
    \label{fig:intro}
\end{figure}

This paper proposes partitioning a DNN between the edge and the host platform to perform coordinated inference to enhance the energy-efficiency (Frames/J) and throughput (Frames/second) of the edge under the channel bandwidth and accuracy constraints [Figure \ref{fig:intro}(c)]. We present a DNN as an information encoding pipeline that can be partitioned between the edge and the host at an intermediate layer. The output feature map of that intermediate layer is further encoded/compressed to reduce the data volume, and transmitted to the host.
We show that from the perspective of the edge device, there is a clear trade-off between the inference and transmission demand depending on how the inference task of a network is allocated. The major contributions of this paper are: 

\begin{itemize}
\item We introduce a partitioned inference approach with feature space encoding, where the edge platform processes inference up to an intermediate layer of the network, and transmit the output features to the host platform for inference of the rest of the network. 

\item Based on the tradeoff analysis of convolutional neural networks (CNNs), we propose a design guideline for partitioning that allocates convolutional layers at the edge and the rest fully-connected layers at the host. 

\item We introduce feature space encoding where the output features are compressed  (loss-less or lossy) before transmission to further enhance the bandwidth utilization. We characterize the accuracy impact of feature space encoding at different layers.

\item We propose split re-training of the DNN to enhance the accuracy with feature space encoding. The partition at the host is fine-tuned by augmenting the training data with encoded features of the intermediate layer, while the partition at the edge remains unchanged. 


\end{itemize}


The advantage of the proposed partitioning approach with feature encoding is demonstrated by the energy/throughput analysis based on an inference engine with an integrated JPEG encoder.
The simulation with AlexNet shows that the JPEG-based lossy feature encoding method reduces the size of the last convolutional layer features of AlexNet by 28x with 1\% accuracy loss. 
Fine tuning of the partitioned network further reduces the transmission demand by 11\% at the same accuracy. We show that the proposed approach improves the system energy efficiency and throughput by 15.3x and 16.5x compared to performing entire inference at the host, and 2.3x and 2.5x compared to performing the entire inference at the edge. It is shown that the improvement depends on the network complexity as well as the transmission bandwidth. Moreover, we show that, although partitioning at the last convolution layer is an efficient design approach, there is a need for dynamic control of the partitioning position to further enhance the throughput or energy.

\section{Related Work}
With recent advance in deep learning techniques, application of DNNs to the IoT environment is being actively investigated.
A typical system configuration uses edge platforms for sensing visual data, which is transmitted to and processed by the host with a DNN inference engine. This configuration is appealing in the applications where the host make central decision and control, such as vehicle detection and recognition \cite{vehicle}, remote monitoring \cite{monitoring}, and scene analysis \cite{crowd}.
Although this approach relieves the inference demand for the edge platforms, its performance will largely depend on reliable transmission of the images through a wireless channel with limited bandwidth.

Recent innovation in network compression \cite{deep_compression} and hardware/ architectural acceleration techniques \cite{eyeriss} has enabled edge platforms with an integrated image sensor and deep learning inference engine \cite{integ_1}. 
When these edge platforms are used to perform the entire inference for delivering the result to the host, large resource demand of deep neural networks will limit their performance.

As opposed to the entire inference at the edge or at the host, a recent study presented a distributed structure of edge devices performing inference of a shallow part of the network \cite{distributed}. However, network partitioning presented in this study was not based on the energy and throughput analysis of the system.  
Moreover, it did not apply an encoding technique to the intermediate-layer features before transmitting them.  
Although a few studies have investigated the accuracy impact of input image encoding \cite{JPEG_encoding} and the weight compression \cite{deep_compression}, there has been no work that explored the effect of intermediate-layer feature encoding on the neural network performance.

In this paper, we build on the prior studies by proposing a design guideline for network partitioning based on the layer-wise trade-off study on the energy-efficiency and throughput of the edge platform under accuracy and bandwidth constraints. The analysis of inference partitioning incorporates the effect of feature encoding in different layers of the network on the transmission demand and the classification accuracy. 

\section{Partitioning of Inference with Feature Space Encoding}

Our system model includes an edge and a host platform, with the edge platform capturing visual data, which is processed through neural network inference, and the output is gathered at the host platform.
Here, the goal is to make the neural network inference results available at the host platform at a target accuracy, with maximum energy efficiency and throughput at the edge device.
To achieve this goal we propose partitioning a network at an intermediate layer, whose features are encoded and transmitted to the host for the rest of the inference.

\subsection{DNN as an Information Encoding Pipeline}


\begin{figure}[t]
  \centering
  \centerline{\includegraphics[trim={5cm 5.6cm 8cm 8cm},clip, width=0.99\linewidth]{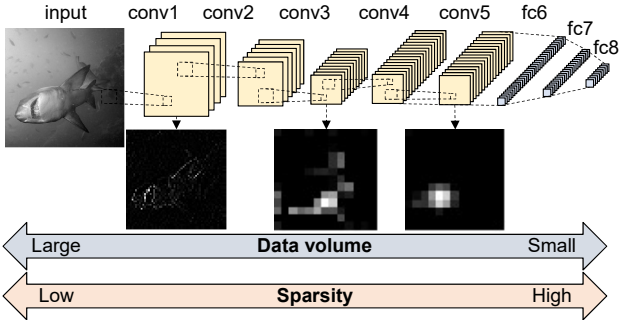}}
  \caption{Visualization of one of the feature maps in different layers of AlexNet, indicating the sparsity increases as the feature is in deeper layers.
}
    \label{fig:feature}
\end{figure}

We view a DNN as an information encoding pipeline. Each layer in a DNN encodes the input features into the output feature space. The size of the output feature space reduces when it passes through a max-pooling operation. Moreover, as each layer encodes the information into a different feature space, the characteristics of the output feature maps such as sparsity or entropy change as well [Figure \ref{fig:feature}]. Therefore, if conventional signal/image encoding and compression techniques are applied at the output feature maps, we can expect different responses at different layers. 

With the preceding view of a DNN as an information encoding pipeline, the edge platform can now be regarded as an encoding engine that processes the raw visual data into a different space before delivering it to the host. The processing involves information extraction through DNN layers and data compression through algorithmic encoding (e.g., JPEG) of the features. The overall compression level is controlled by the parameters, namely, how many layers are processed at the edge (partitioning position) and quality factor ($QF$) of the algorithmic encoder. 


\subsection{Edge-Host Partitioning of Inference}
\begin{figure}[t]
  \centering
  \centerline{\includegraphics[trim={5cm 2.8cm 5cm 5cm},clip, width=0.89\linewidth]{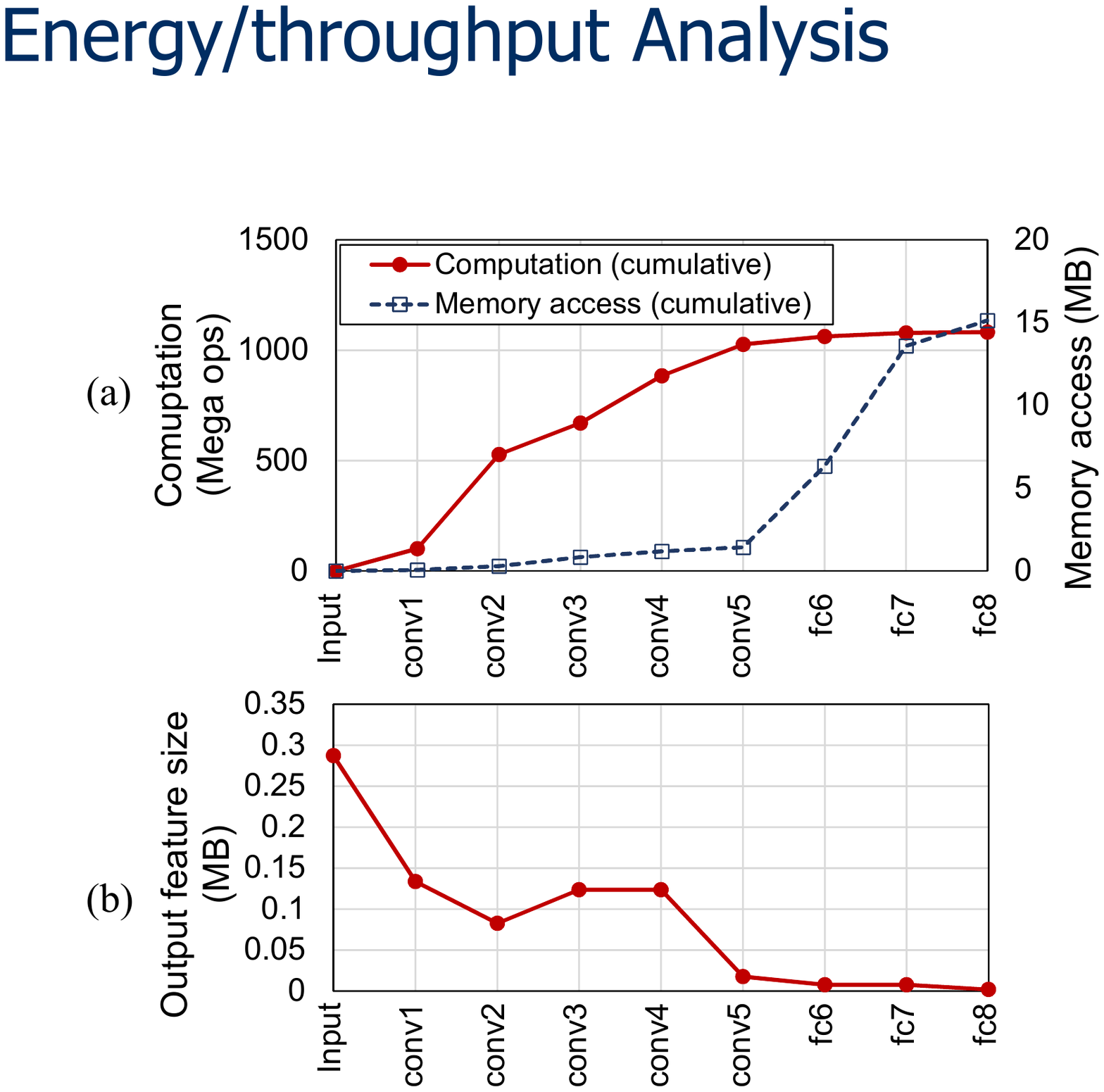}}
  \caption{ (a) Cumulative computation and memory access (for compressed weights) demand of inference and (b) output feature size for each layer of AlexNet.
}
    \label{fig:demand}
\end{figure}
We first demonstrate the tradeoff analysis of the partitioning approach based on the inference/transmission demand of AlexNet when it is partitioned at different layers [Figure \ref{fig:demand}].
When the entire inference is performed at the host platform, the edge platform should transmit the input images to the host.
Figure \ref{fig:demand}(b) shows that the input feature size for AlexNet is nearly 0.3 MB with 16-bit precision. To satisfy a typical frame rate of 30 frames/sec, the required bandwidth can go up to 70 Mbps, which is not generally accommodated by conventional low-power transmitters. 
To mitigate the transmission overhead, the edge platform can process the whole network inference and transmit the output of the network. 
However, the entire network inference requires huge computation and memory demand, as illustrated in Figure \ref{fig:demand}(a).

By allowing transmission of the intermediate features, the inference partitioning approach enables the layer-wise tradeoff for higher energy-efficiency and throughput at the edge.
The system performance with inference partitioning will be determined depending on which layer the network is partitioned.
If we partition the network at a deeper layer of a CNN by including fully-connected layers at the edge, the system throughput and energy efficiency will degrade mainly because of the memory access demand of fully connected layers. 
If we partition the network at an earlier convolutional layer, transmission demand will be still limiting the system performance because of its larger feature size.

Based on these observations, we propose an initial design guideline that partitions the network at the end of convolution layers.
An edge device designed to perform convolutional layer inference will be useful because a set of convolutional layers is widely used as a feature extractor for backbone networks such as Convolutional-RNN.
Another benefit of this partitioning approach is that the data flow of the edge platform can be optimized for convolutional layers instead of considering the heterogeneous data flow in fully-connected layers. 
Also, smaller kernel size of the convolutional layers makes them easily fit into the edge devices with limited storage.

\subsection{Feature Space Encoding}

\begin{figure}[t]
  \centering
  \centerline{\includegraphics[trim={5.3cm 3.5cm 6cm 5.5cm},clip, width=0.89\linewidth]{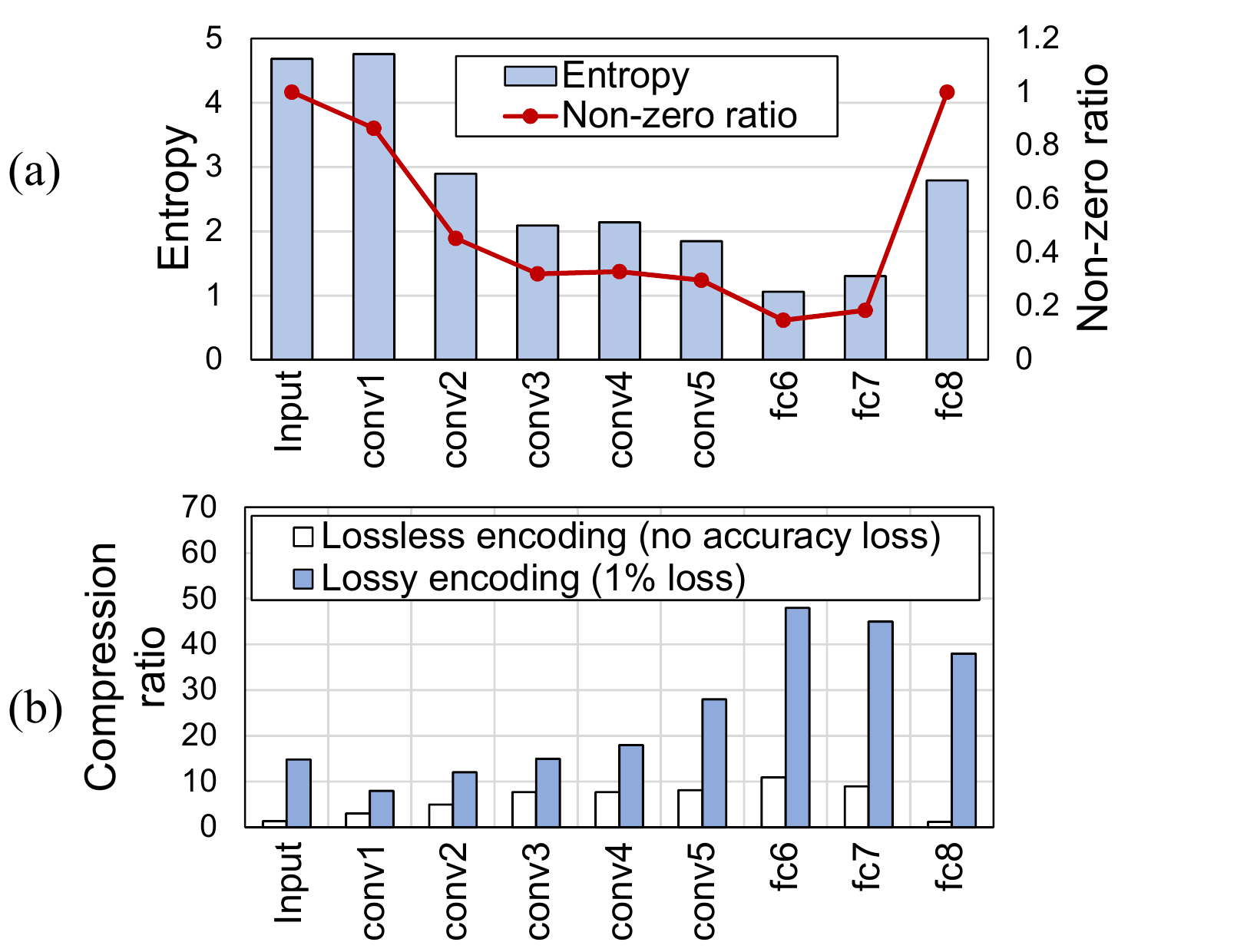}}
  \caption{ (a) Entropy and non-zero ratio of features in different layers in AlexNet. (b) Compression ratio of each layer features with lossless and lossy encoding.
}
    \label{fig:encoding}
\end{figure}

Inference partitioning at an intermediate layer implies that its output feature maps have to be transmitted to the host side. 
To further improve energy-efficiency and throughput, we propose encoding the feature maps before transmission. 

A primary reason for encoding the intermediate features is to reduce the transmission demand.
Moreover, data representation of the original feature maps is generally sparse, making them easily compressed.
Figure \ref{fig:encoding}(a) shows that the sparsity increases as the features are in deeper layers, with lower non-zero ratio and entropy.  
However, the feature encoding may cause loss of information, which leads to the degradation of the inference accuracy. 
Therefore, to leverage the benefit of feature encoding while minimizing the loss of accuracy, we examine the robustness of each layer features to different encoding methods including lossless and lossy encoding.   

\subsubsection*{\textbf{Lossless Encoding}}

Sparse representation of the feature maps is appealing property for lossless encoding.
We apply a common lossless encoding method, run-length encoding combined with the Huffman encoding, which is the last part of the JPEG encoder pipeline. 
As this method converts sequences or zeros into the pre-defined codes, higher ratio of zeros in deeper layer features leads to higher compression ratio, as illustrated in Figure \ref{fig:encoding}(b). 
Although the encoded features are perfectly reconstructed at the host, resulting in no accuracy loss, its compression ratio is limited to 3-10x.

\subsubsection*{\textbf{Lossy Encoding}} 
We apply JPEG encoding (lossy) to the intermediate feature, to achieve further compression of the features at the expense of the loss of accuracy.
Figure \ref{fig:encoding}(b) show the compression ratio of each layer features obtained at the 1\% accuracy loss. 
As lower entropy leads to higher performance of lossy encoding, deeper layer features with lower entropy generally shows higher compression ratio. 
With 1\% accuracy loss, lossy encoding achieves 5-50x compression, which significantly reduces the demand on transmission bandwidth and energy. 

\subsubsection*{\textbf{Fine-Tuning of the Partitioned Network}} 
\begin{figure}[t]
  \centering
  \centerline{\includegraphics[trim={4.5cm 3.3cm 5cm 5.5cm},clip, width=0.99\linewidth]{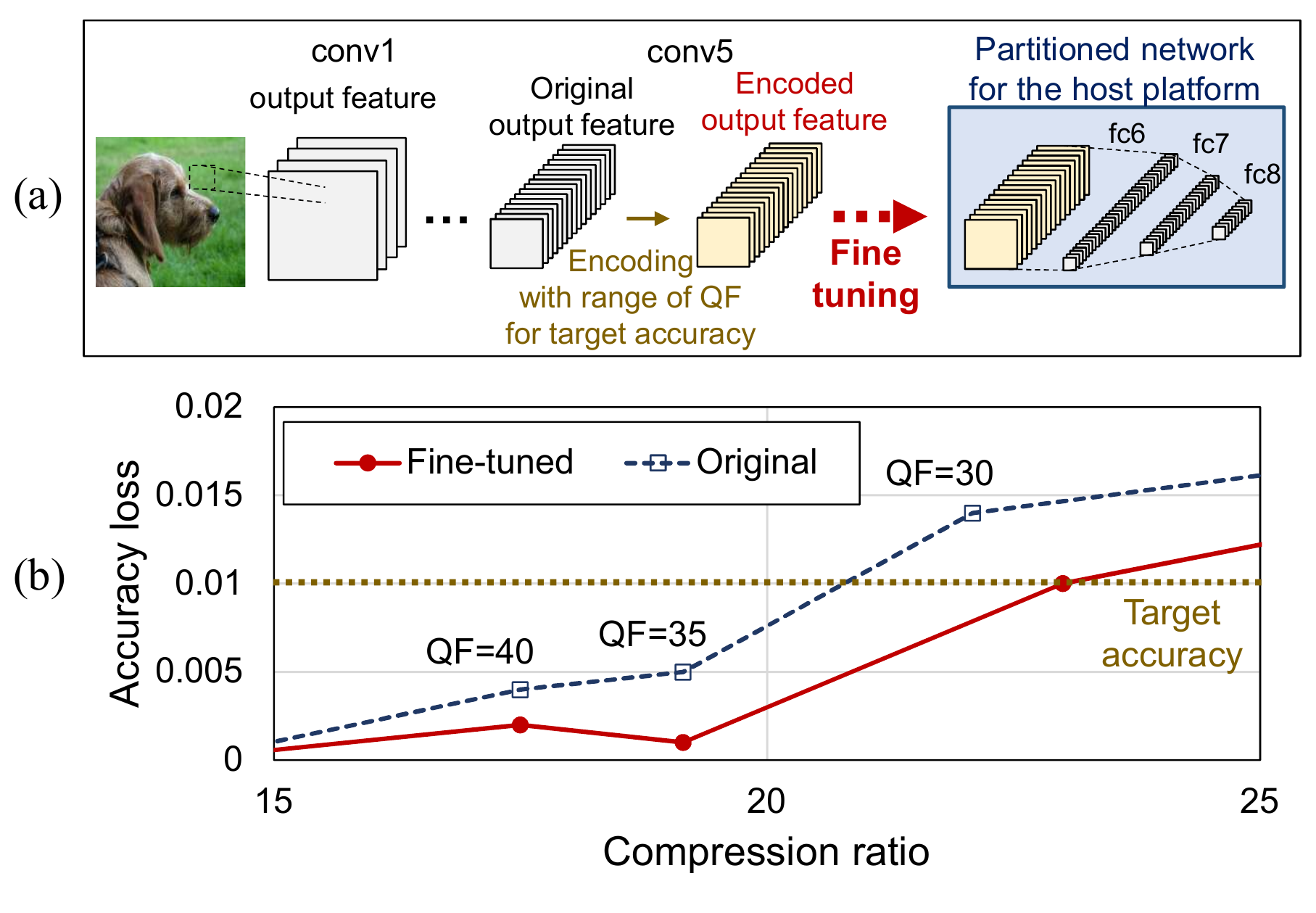}}
  \caption{ (a) Process of fine tuning with encoded intermediate features. (b) Compression ratio and accuracy loss of a partitioned network according to various QF values used to encode the conv5 output features.
}
    \label{fig:tuning}
\end{figure}
We propose to apply a fine-tuning technique to enhance the accuracy under lossy encoding of the features. In other words, improved accuracy due to fine-tuning will allow higher compression for the same accuracy loss. 
Fine tuning is commonly used technique to enhance the accuracy of the entire network using augmentation of input space. However, since the transformation is applied at an intermediate layer, we propose split re-training of the network, where \textit{only the network partition at host is re-trained}, while the edge partition remains same [Figure \ref{fig:tuning}(a)].
The augmentation of the input to the partitioned network is determined by the compression performance of the original features of the target layer.
Figure \ref{fig:tuning}(b) shows an example of the AlexNet conv5 layer, where the target accuracy (1\% loss) with the original network is achieved with the quality factor (QF) between 30 and 35.
Therefore, we prepare the input features for fine tuning by encoding the intermediate features of the training dataset with the randomly selected QF between 30 and 35.
The encoded features are then supplied to fine-tune the network partition for the host platform. 
As Figure \ref{fig:tuning}(b) shows, fine tuning further enhances the compression ratio by 11\% at 1\% accuracy loss.


\section{Simulation Results}
\subsection{Inference Engine Design and Modeling}
\begin{figure}[t]
  \centering
  \centerline{\includegraphics[trim={4.3cm 7cm 4cm 6.5cm},clip, width=0.99\linewidth]{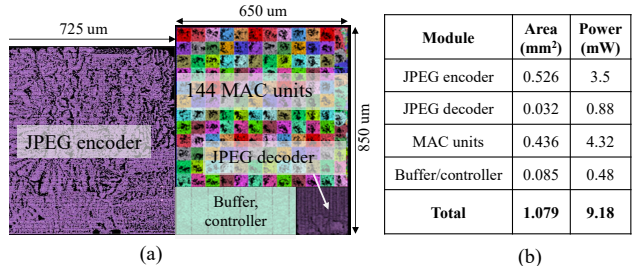}}
  \caption{ Hardware design of an inference engine. (a) Layout of the synthesized design and (b) area and power breakdown of the system.
}
    \label{fig:hardware}
\end{figure}

We designed an inference engine that performs inference and feature encoding to enable the energy and throughput analysis of the proposed approach. 
The engine includes an array of 144 16-bit MAC units, a JPEG encoder for feature space encoding, and an on-chip buffer to store the input/output feature maps [Figure \ref{fig:hardware}(a)]. 
As we assume the weights are stored in a compressed format through the JPEG encoding method presented in \cite{date}, we included a JPEG decoder for decoding the compressed weights.
The design was synthesized into an ASIC with a 28nm process, and its area and power characteristics are shown in Figure \ref{fig:hardware}(b)

For the weight storage, we use an off-chip LPDDR DRAM with 640 pJ/32 bit data access and 12.8 MB/s bandwidth \cite{jedec} \cite{EIE}. 
Feature transmission is performed by an ultra low power 2.4GHz 802.11bgn transmitter module \cite{nlink} that has 1, 2, and 22 Mbps datarate modes with 62.7, 99, and 660 mW power consumption, respectively (we use a 2 Mbps mode as a default).
As we model the processing engine of 2-stage pipeline with inference and feature encoding/transmission stages, the system throughput is determined by the largest latency among these two stages.

\subsection{Energy/Throughput Analysis}


\subsubsection*{\textbf{Analysis without Feature Space Encoding}}
\begin{figure}[t]
  \centering
  \centerline{\includegraphics[trim={4.7cm 2.9cm 5.5cm 5.7cm},clip, width=0.89\linewidth]{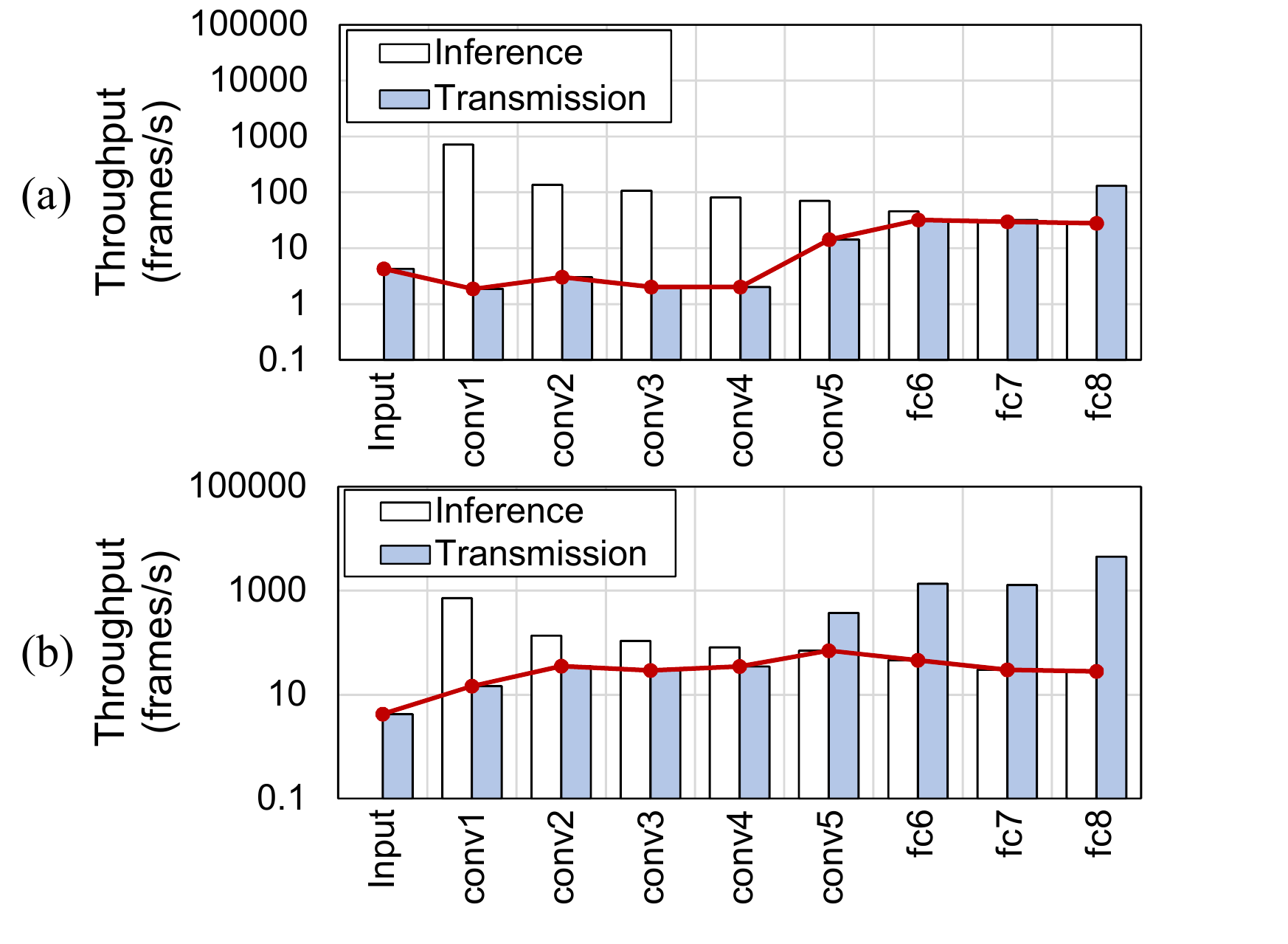}}
  \caption{ System throughput of the edge platform. (a) Without encoding and (b) with lossy feature encoding (at the accuracy loss=1\%) and weight compression. Here, the encoding latency is included in the transmission latency.
}
    \label{fig:throughput}
\end{figure}

\begin{figure}[t]
  \centering
  \centerline{\includegraphics[trim={5cm 1.5cm 5.5cm 5.1cm},clip, width=0.89\linewidth]{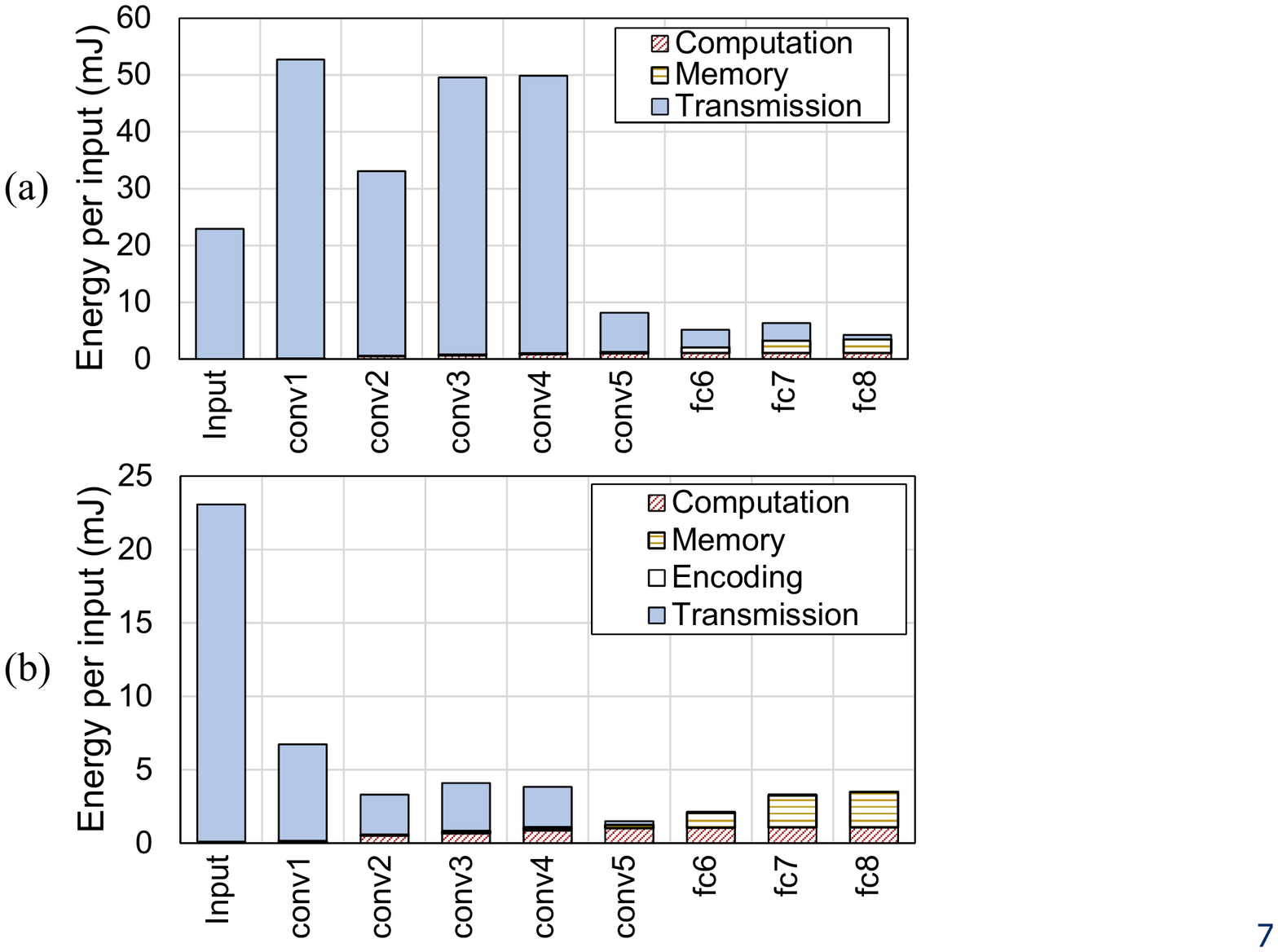}}
  \caption{ System energy breakdown of the edge platform. (a) Without encoding and (b) with lossy feature encoding (at the accuracy loss=1\%) and weight compression.
}
    \label{fig:energy}
\end{figure}

Figure \ref{fig:throughput}(a) and \ref{fig:energy}(a) show the system energy/throughput when AlexNet is partitioned at different layers.
For inference at the host (i.e., input feature transmission from edge), we assume the edge applies JPEG encoding on the raw image (input feature). The quality factor for the input compression is estimated by considering 1\% accuracy loss at the classification.
Even with 15x compression of the input features, the energy/throughput performance of the host inference approach is still limited by transmission. 
Next, we consider processing a set of CNN layers at the edge. Here, we assume no feature encoding at intermediate layers.
When the network is partitioned at a deeper layer, inference becomes the system bottleneck. 

By including only a part of the network at the edge, the inference partitioning approach enhances the system performance.
The maximum throughput and minimum energy of the edge platform is realized when the network is partitioned after the first fully-connected layer (fc6). At this point, the energy-efficiency and throughput is enhanced by 1.2x and 1.15x compared to the entire inference at the edge, and 4.5x and 7.5x compared to the entire inference at the host, respectively. 
 

\subsubsection*{\textbf{Analysis with Feature Space Encoding}}

When the network is partitioned at convolutional layers, their large feature sizes still make the transmission become the throughput and energy bottleneck.
Figure \ref{fig:throughput}(b) and \ref{fig:energy}(b) show that the feature encoding approach reduces transmission latency and energy consumption with negligible encoding overhead.
The improved transmission throughput enhances the system throughput in convolutional layers where transmission was the bottleneck in the previous case without feature encoding.
Feature encoding significantly enhances the system energy as well when the network is partitioned at convolutional layers. 

With feature encoding, the optimal partitioning layer for maximum throughput and minimum energy becomes the last convolutional layer (conv5). 
If the network is partitioned at fully-connected layers, the benefit of feature encoding is limited because their original feature sizes are relatively small and the system energy is dominated by inference.
If we partition the network at an earlier layer, transmission is still the bottleneck because of their larger feature size.
In these layers, the transmission throughput can be improved by compressing the features further, but further compression will result in more accuracy loss. 

\begin{figure}[t]
  \centering
  \centerline{\includegraphics[trim={3cm 6.7cm 10cm 5.5cm},clip, width=0.90\linewidth]{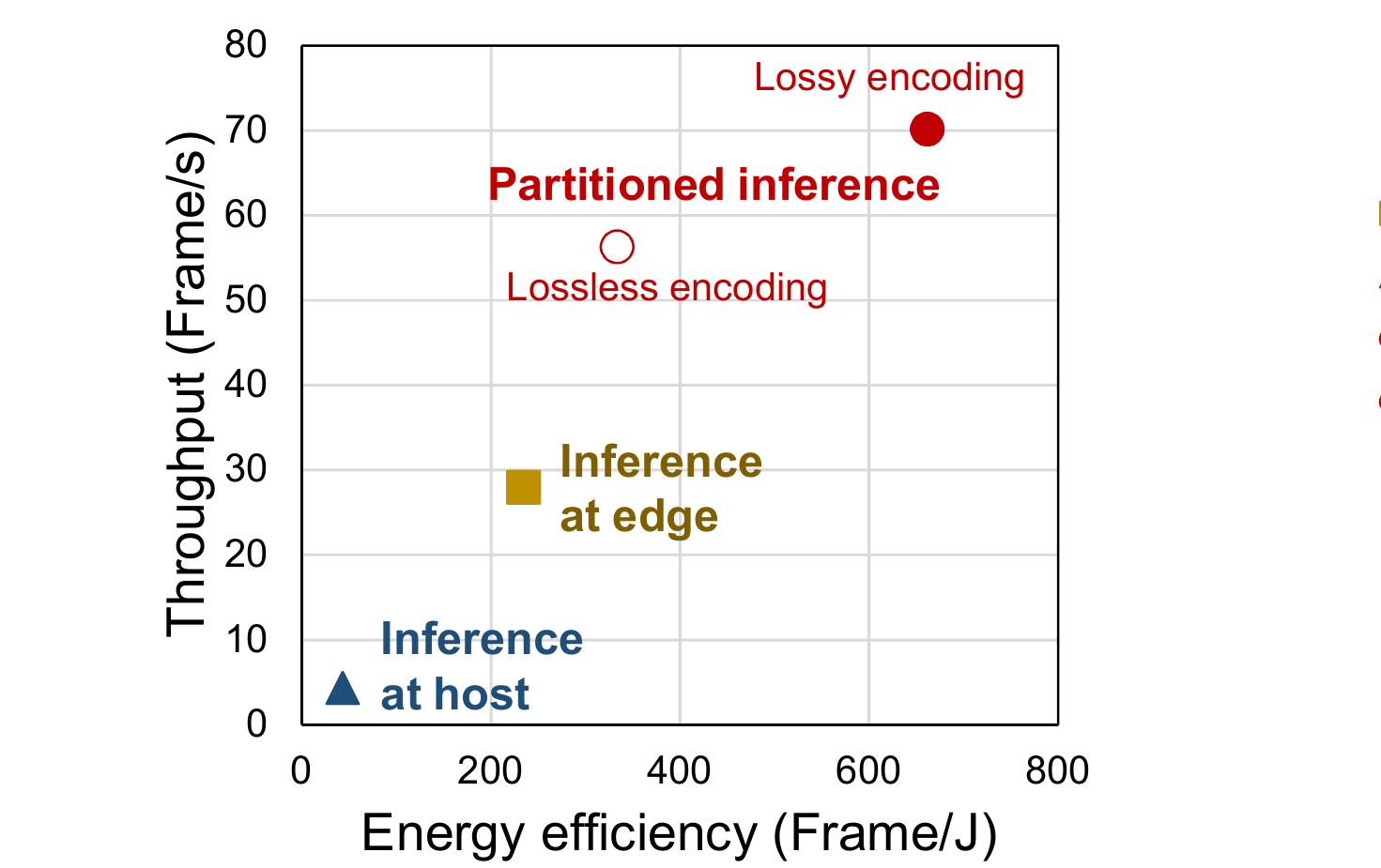}}
  \caption{  Summary of improvement of the throughput and energy-efficiency of the proposed partitioning with feature encoding.
}
    \label{fig:summary}
\end{figure}

Figure \ref{fig:summary} illustrates the summary of performance improvement obtained by the proposed partitioned inference with feature encoding on AlexNet.
Higher compression of the features through lossy encoding combined with fine tuning provides more gain in energy efficiency and throughput.
The proposed partitioning approach with fine-tuning assisted feature encoding achieves energy reduction and throughput improvement by 15.3x and 16.5x than the entire inference at the host, and 2.3x and 2.5x than the entire inference at the edge, respectively.


\subsection{Discussions}
 \subsubsection*{\textbf{Effect of Network Types}}

\begin{figure}[t]
  \centering
 \centerline{\includegraphics[trim={5cm 1.9cm 5cm 6.7cm},clip, width=0.99\linewidth]{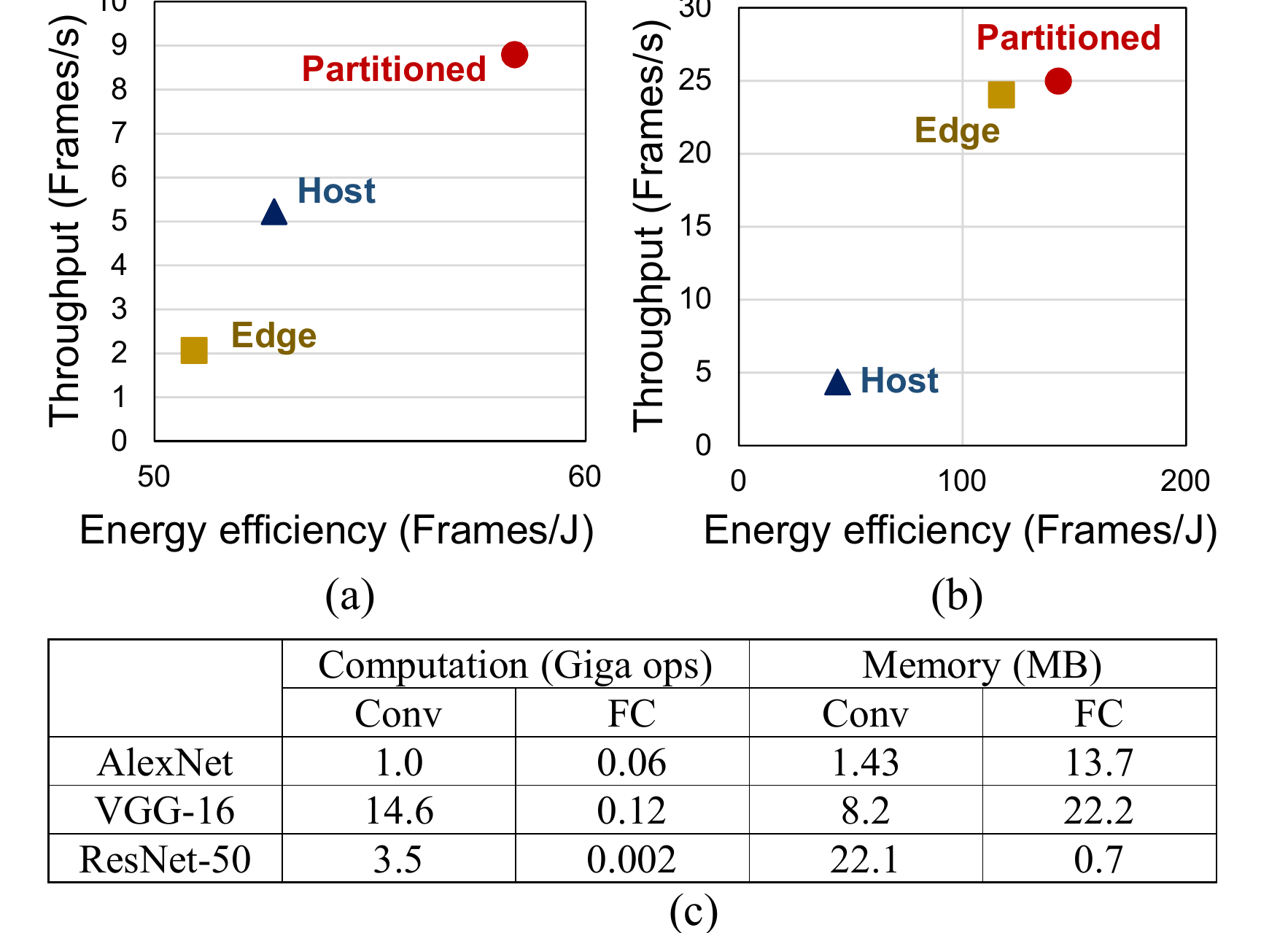}}
  \caption{Throughput energy efficiency of each approach for (a) VGG-16 and (b) ResNet-50. (c) Inference demand of each network (assuming compressed weights).
}
    \label{fig:network}
\end{figure}

The peformance of the proposed inference partitioning approach will vary depending on the type of the neural network. We perform the simulation with two other network models, VGG-16 \cite{vgg} and ResNet-50 \cite{resnet} [Figure \ref{fig:network}].
In VGG-16, large memory access demand of fully-connected layers significantly degrades the performance of the edge inference approach, compared to the AlexNet case. 
By including only convolutional layers at the edge, the proposed partitioning approach can avoid large memory access of fully-connected layers, achieving 1.2x and 4.3x higher energy efficiency and throughput than the edge inference model.  

ResNet-50 has deep convolutional layers followed by one fully-connected layer, with smaller feature sizes making its computation demand smaller than VGG-16. 
Therefore, instead of inference at the host, processing the entire inference at the edge leads to higher throughput and energy efficiency. 
The proposed partitioning approach provides better efficiency by removing the last layer inference from the edge, since it can avoid large memory demand of the fully-connected layer.


 \subsubsection*{\textbf{Effect of Transmission Channel}}

\begin{figure}[t]
  \centering
  \centerline{\includegraphics[trim={4.5cm 6.8cm 5cm 5.4cm},clip, width=0.99\linewidth]{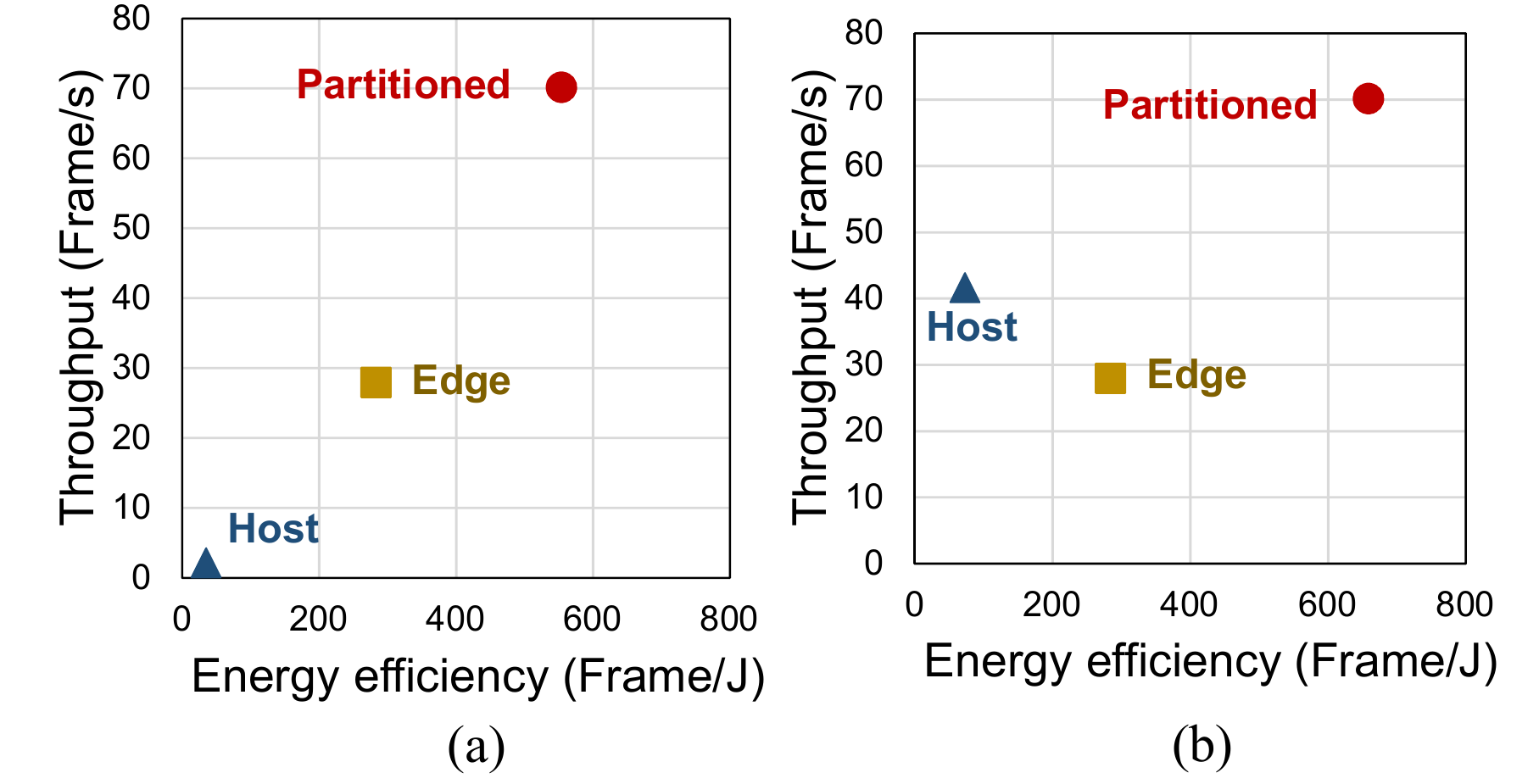}}
  \caption{ Throughput energy efficiency of each approach for (a) Low BW (1 Mbps) and (b) high BW (22 Mbps) channel.
}
    \label{fig:BW}
\end{figure}

Figure \ref{fig:BW} shows the performance of each approach with two different transmission channels (low bandwidth=1 Mbps, high bandwidth=22 Mbps) on AlexNet. 
Even with a transmission bandwidth increase, the performances of the edge inference and the partitioned inference approaches remain almost same because they are not transmission-bottlenecked. On the other hand, host inference approach enhances the system throughput with reduced transmission latency of the input images. We observe that the convolutional layer partitioning approach provides performance gain over the edge or host inference approaches. 

\begin{figure}[t]
  \centering
  \centerline{\includegraphics[trim={4.5cm 4.3cm 6.8cm 6.4cm},clip, width=0.89\linewidth]{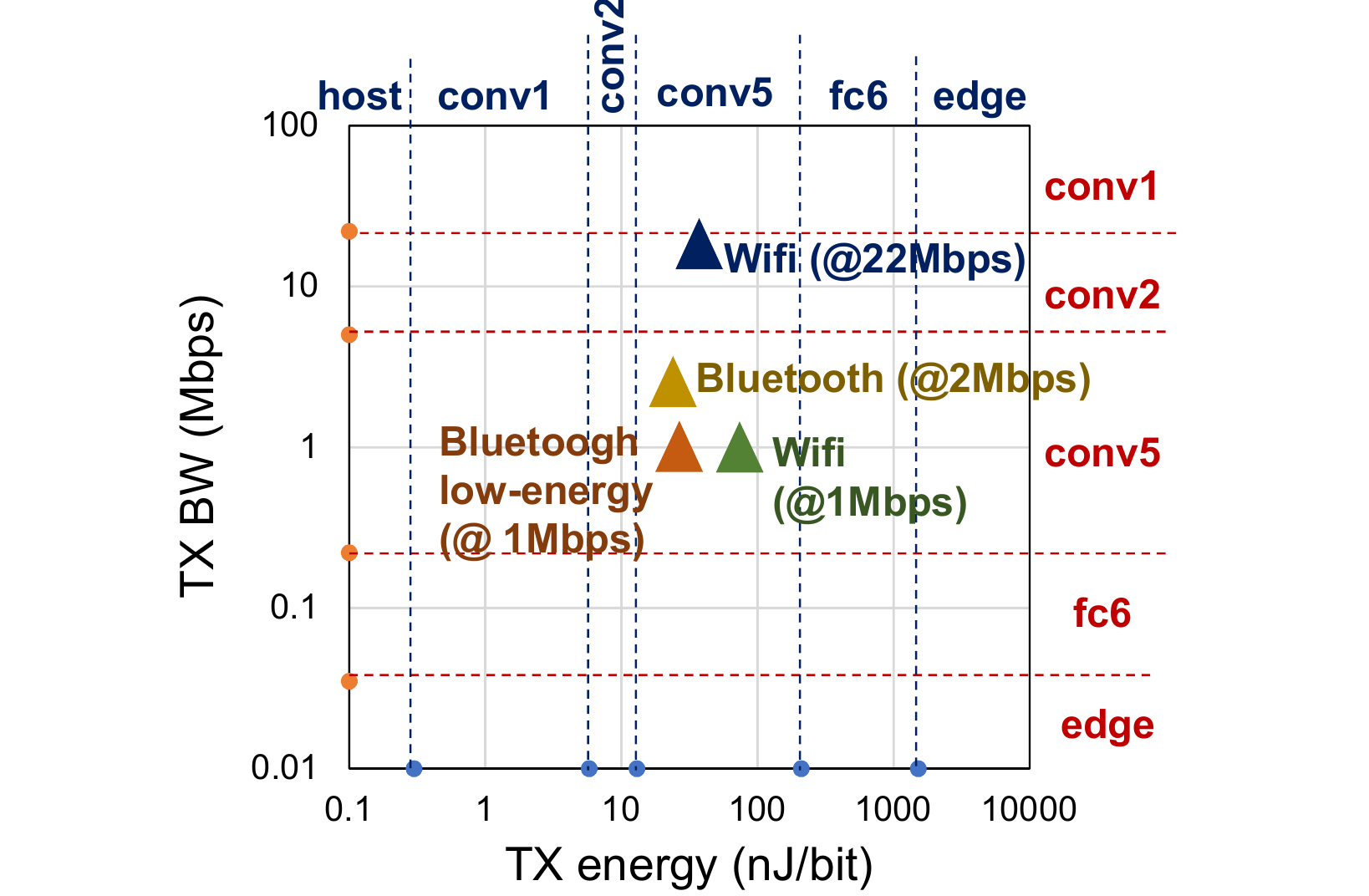}}
  \caption{ The optimal partitioning layer in terms of the throughput (y-axis) and energy-efficiency (x-axis) according to the transmission BW and energy of a transmitter.
}
    \label{fig:BW_optimal}
\end{figure}

We further characterize the effect of transmission channel on the network partitioning. 
Figure \ref{fig:BW_optimal} shows how the optimal partitioning layer is shifted according to the transmission bandwidth and energy consumption of a transmitter. We observe that in a large-bandwidth case, the optimal partitioning layer for maximum throughput is shifted to the second layer since transmission can accommodate larger features. However, with a lower channel bandwidth, it is preferred to process more layers by partitioning at deeper layers. 
Although convolutional layer partitioning continues to provide effective gain in a wide range of commercial low-power transmitter models, we believe the future work can include designing a controller that dynamically adapts to the variation of transmission channel by shifting the partitioning layer, together with controlling the data volume at a certain layer.


\section{Conclusions}
In this paper, we introduced partitioning a DNN inference task between the edge and the host platforms to improve the energy-efficiency and throughput of the edge device.
The improvement is leveraged by encoding the features of an intermediate layer, with the help of fine tuning of the partitioned network. We demonstrated that the proposed partitioning coupled with feature encoding significantly enhances the energy-efficiency and throughput of the edge compared to the entire inference processing at the edge of at the host platform. 

\bibliographystyle{unsrt}
\bibliography{references} 


\end{document}